\documentclass[10pt,twocolumn,letterpaper]{article}

\usepackage{cvpr} \usepackage{times} \usepackage{epsfig}
\usepackage{graphicx} \usepackage{amsmath} \usepackage{amssymb}
\usepackage{mathtools} \usepackage{enumitem,kantlipsum}
\usepackage{cprotect} \usepackage{multirow} \usepackage{collcell}
\usepackage{subfig} \usepackage{float} \usepackage{color}
\usepackage[toc,page]{appendix}
\usepackage[pagebackref=true,breaklinks=true,letterpaper=true,colorlinks,bookmarks=false]{hyperref}

\def\mhyphen{{\hbox{-}}}

\cvprfinalcopy 

\ifcvprfinal\fi
\begin{document}

\title{Video Scene Parsing with Predictive Feature Learning}

\author{Xiaojie Jin$^1$ \ Xin Li$^2$ \ Huaxin Xiao$^2$ \ Xiaohui Shen$^3$ \ Zhe Lin$^3$ \ Jimei Yang$^3$ \\ Yunpeng Chen$^2$ \ Jian Dong$^4$ \ Luoqi Liu$^4$ \ Zequn Jie$^2$ 　 Jiashi Feng$^2$ \ Shuicheng Yan$^{4,2}$\\
  \small $^1$NUS Graduate School for Integrative Science and Engineering, NUS\\
  \small$^2$Department of ECE, NUS \qquad $^3$Adobe Research \quad $^4$360 AI
  Institute}

\maketitle

\begin{abstract}
  In this work, we address the challenging video scene parsing problem by
  developing effective representation learning methods given limited parsing
  annotations. In particular, we contribute two novel methods that constitute a
  unified parsing framework. (1) {\textbf{Predictive feature learning}} from
  nearly unlimited unlabeled video data. Different from existing methods
  learning features from single frame parsing, we learn spatiotemporal
  discriminative features by enforcing a parsing network to predict future
  frames and their parsing maps (if available) given only historical frames. In
  this way, the network can effectively learn to capture video dynamics and
  temporal context, which are critical clues for video scene parsing, without
  requiring extra manual annotations.  (2) {\textbf{Prediction steering
      parsing}} architecture that effectively adapts the learned spatiotemporal
  features to scene parsing tasks and provides strong guidance for any
  off-the-shelf parsing model to achieve better video scene parsing
  performance. Extensive experiments over two challenging datasets, Cityscapes
  and Camvid, have demonstrated the effectiveness of our methods by showing
  significant improvement over well-established baselines.
\end{abstract}

\begin{figure*}[t!]
\captionsetup[subfigure]{labelformat=empty}
	\centering
	\subfloat[]{%
	    \includegraphics[width=\linewidth]{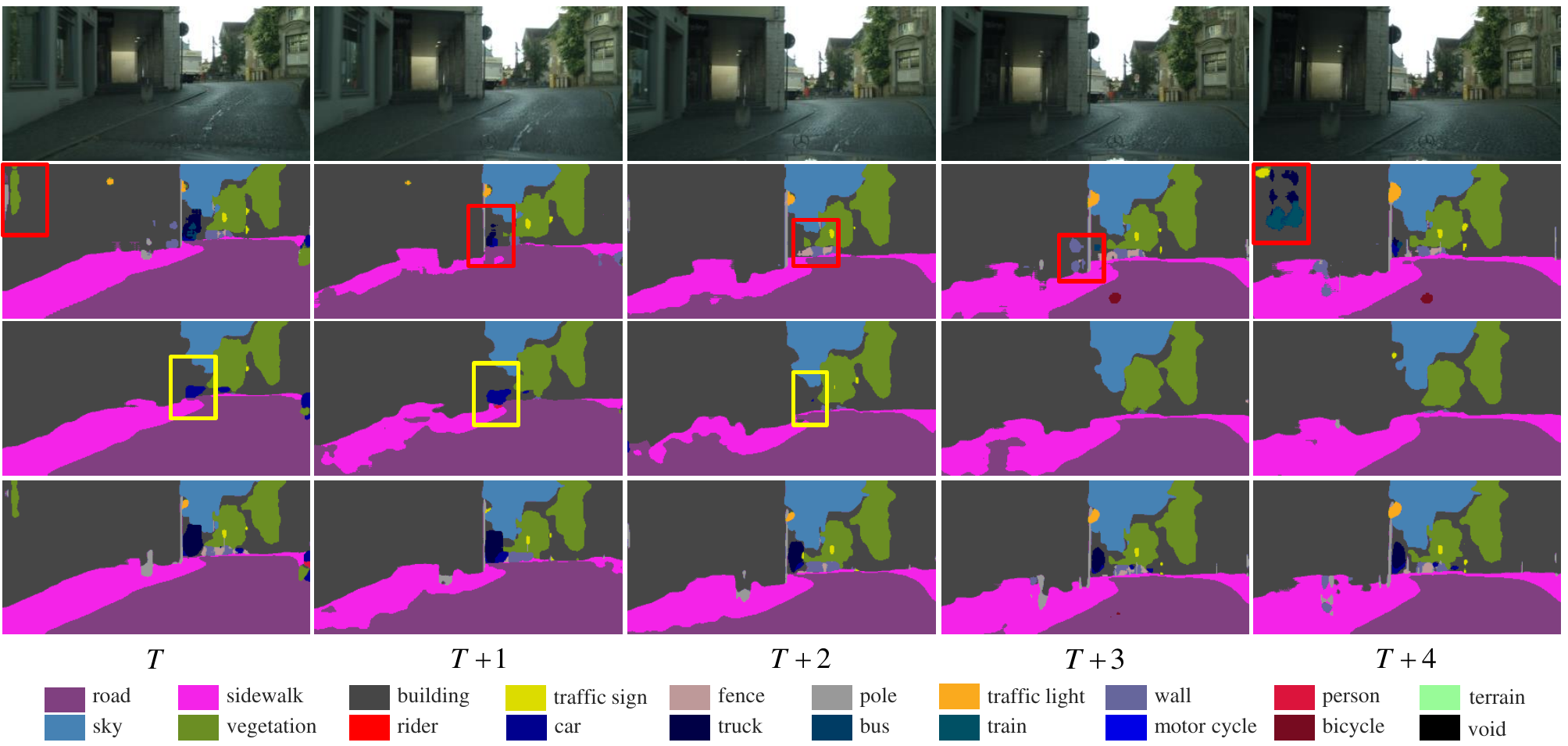}%
	  }
	\vspace{-2mm}
	\caption{Illustration on core ideas of predictive feature learning for
          video scene parsing. \textbf{Top}: a five-frame sequence from the
          Cityscapes video dataset. \textbf{Second row}: frame parsing produced
          by the conventional VGG16 model~\cite{chen2014semantic}. Since it is
          unable to model temporal context, severe noise and inconsistency
          across frames can be observed within the red boxes. \textbf{Third
            row}: results from predictive feature learning and parsing. The
          regions showing inconsistency in the second row are classified
          consistently across frames by predictive feature learning. Besides,
          the motion trajectories of cars are correctly captured (see yellow
          boxes). \textbf{Bottom}: parsing maps produced by PEARL with better
          accuracy and temporal consistency. PEARL combines the advantages of
          traditional image parsing model (the second row) and predictive
          parsing model (the third row). Best viewed in color and zoomed pdf.}
          \label{fig:vis_cmp}
\end{figure*}
\section{Introduction}
Video scene parsing (VSP) aims to predict per-pixel semantic labels for every
frame in scene videos recorded in unconstrained environments. It has drawn
increasing attention as it benefits many important applications like drones
navigation, autonomous driving and virtual reality.

In recent years, remarkable success has been made by deep convolutional neural
network (CNN) models in image parsing tasks
\cite{chen2014semantic,farabet2013learning,fullyconvseg,roy2014scene,schwing2015fully,socher2011parsing,zhang2012efficient,zheng2015conditional}. Some
of those CNN models are thus proposed to be used for parsing scene videos frame
by frame. However, as illustrated in Figure~\ref{fig:vis_cmp}, naively applying
those methods suffers from noisy and inconsistent labeling results across
frames, since the important temporal context cues are ignored. For example, in
the second row of Figure~\ref{fig:vis_cmp}, the top-left region of class
\emph{building} in the frame $T$+4 is incorrectly classified as \emph{car},
which is temporally inconsistent with the parsing result of its preceding
frames. Besides, for current data-hungry CNN models, finely annotated video data
are rather limited as collecting pixel-level annotation for long videos is very
labor-intensive.  Even in the very recent scene parsing dataset Cityscapes
\cite{cityscape}, there are only 2,975 finely annotated training samples
\emph{vs.}\ overall 180K video frames. Deep CNN models are prone to over-fitting
the small training data and thus generalize badly in real applications.

To tackle these two problems, we propose a novel \textbf{P}arsing with
pr\textbf{E}dictive fe\textbf{A}tu\textbf{R}e \textbf{L}earning (\textbf{PEARL})
approach which is both annotation-efficient and effective for VSP.  By enforcing
them to predict future frames based on historical ones, our approach guides CNNs
to learn powerful spatiotemporal features that implicitly capture video dynamics
as well as high-level context like structures and motions of
objects. Attractively, such a learning process is nearly annotation-free as it
can be performed using any unlabeled videos. After this, our approach further
adaptively integrates the obtained temporal-aware CNN to steer any image scene
paring models to learn more spatial-temporally discriminative frame
representations and thus enhance video scene parsing performance substantially.

Concretely, there are two novel components in our proposed approach:
\textit{predictive feature learning} and \textit{prediction steering
  parsing}. As shown in Figure \ref{fig:vis_cmp}, given frames $T$ to
$T\text{+}3$, predictive feature learning aims to learn discriminative
spatiotemporal features by enforcing a CNN model to predict the future frame
$T\text{+}4$ as well as the parsing map of $T\text{+}4$ if available. Such
predictive learning enables the CNN model to learn features capturing the
cross-frame object structures, motions and other temporal cues, and provide
better video parsing results, as demonstrated in the third row of
Figure~\ref{fig:vis_cmp}. To further adapt the obtained CNN along with its
learned features to the parsing task, our approach introduces a prediction
steering parsing architecture. Within this architecture, the temporal-aware CNN
(trained by frame prediction) is utilized to guide an image-parsing CNN model to
parse the current frame by providing temporal cues implicitly. The two parsing
networks are jointly trained end-to-end and provide parsing results with strong
cross-frame consistency and richer local details (as shown in the bottom row of
Figure \ref{fig:vis_cmp}).

We conduct extensive experiments over two challenging datasets and compare our
approach with strong baselines, \textit{i.e.}, state-of-the-art VGG16~\cite{vgg}
and Res101~\cite{residual} based parsing models. Our approach achieves the best
results on both datasets. In the comparative study, we demonstrate its
superiority to other methods that model temporal context, \textit{e.g.}, using
optical flow \cite{epicflow}.

To summarize, we make the following contributions to video scene parsing:
\begin{itemize}
	\setlength\itemsep{0em}
      \item A novel predictive feature learning method is proposed to learn the
        spatiotemporal features and high-level context from a large amount of
        unlabeled video data.
      \item An effective prediction steering parsing architecture is presented
        which utilizes the temporal consistent features to produce temporally
        smooth and structure preserving parsing maps.
      \item Our approach achieves state-of-the-art performance on two
        challenging datasets, \emph{i.e.}, Cityscapes and Camvid.
\end{itemize}

\section{Related Work}
Recent image scene parsing progress is mostly stimulated by various new CNN
architectures, including the fully convolutional architecture (FCN) with
multi-scale or larger receptive fields
\cite{farabet2013learning,fullyconvseg,socher2011parsing} and the combination of
CNN with graphical
models~\cite{chen2014semantic,roy2014scene,zhang2012efficient,zheng2015conditional,schwing2015fully}. There
are also some recurrent neural networks based
models~\cite{mpf,RCNN,pinheiro2013recurrent,dag,reseg}. However, without
incorporating the temporal information when directly applying them to every
frame of videos, the parsing results commonly lack cross-frame consistency and
the quality is not good.

To utilize temporal consistency across frames, the motion and structure features
in 3D data are employed
by~\cite{joint23d,sturgess2009combining,zhang2010semantic}. In addition,
\cite{multi-class,featureregular,RTDF,budgeted_crf} use CRF to model
spatiotemporal context. However, those methods suffer from high computation cost
as they need to perform expensive inference of CRF. Some other methods employ
optical flow to capture the temporal consistency explored
in~\cite{jointof,sevilla2016optical}. Different from above works that heavily
depend on labeled data for supervised learning, our proposed approach takes
advantage of both the labeled and unlabeled video sequences to learn richer
temporal context information.

Generative adversarial networks were firstly introduced in~\cite{goodfellowgan}
to generate natural images from random noises, and have been widely used in many
fields including image synthesis \cite{goodfellowgan}, frame prediction
\cite{lotter2015unsupervised,mathieu} and semantic
inpainting~\cite{trevor16inpainting}.  Our approach also uses adversarial loss
to learn more robust spatiotemporal features in frame predictions. Our approach
is more related to \cite{lotter2015unsupervised,mathieu} by using adversarial
training for frame prediction. However, different
from~\cite{lotter2015unsupervised,mathieu}, PEARL tackles the VSP problem by
utilizing spatiotemporal features learned in frame prediction.

\section{Predictive Feature Learning for VSP}
\subsection{Motivation and Notations}
\label{sec:notations}
The proposed approach is motivated by two challenging problems of video scene
parsing: first, how to leverage the temporal context information to enforce
cross-frame smoothness and produce structure preserving parsing results; second,
how to build effective parsing models even in presence of insufficient training
data.

\begin{figure}
  \subfloat[The framework of predictive feature learning in
  PEARL\label{fig:framework_PFL}]    {\includegraphics[width=\linewidth]{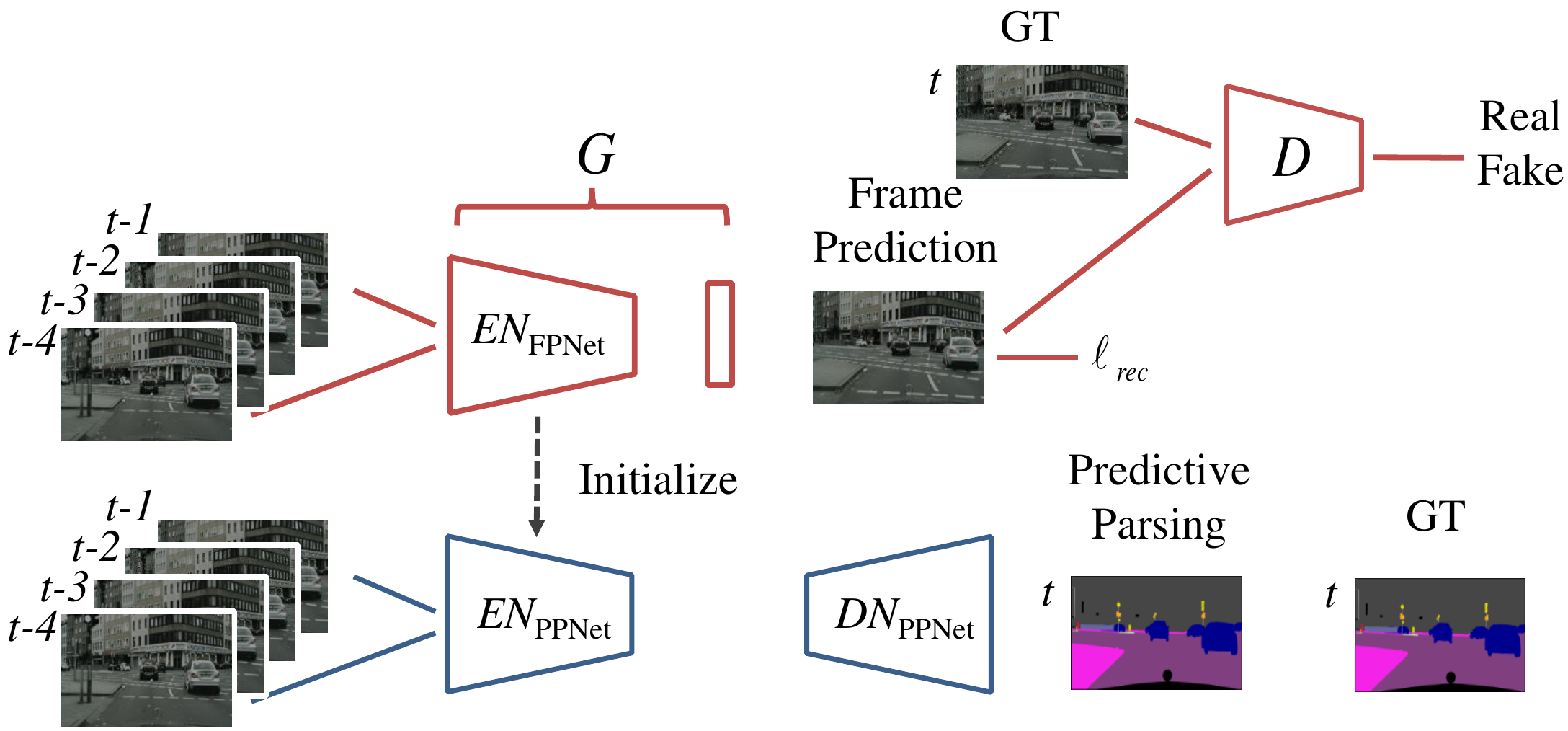}}\\
  \subfloat[The architecture of prediction steering parsing network
  in　PEARL\label{fig:framework_PSP}]
  {\includegraphics[width=\linewidth]{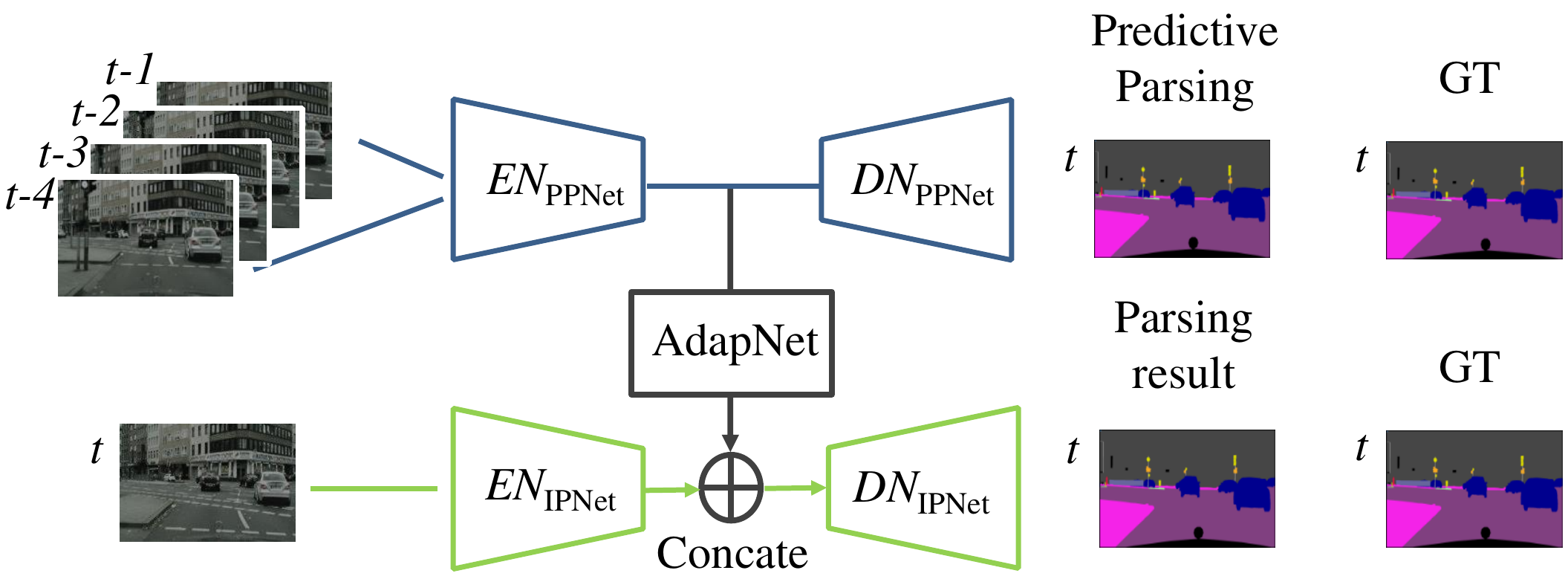}}\\
  \subfloat[A variant of PEARL\label{fig:framework_variant}]
  {\includegraphics[width=\linewidth]{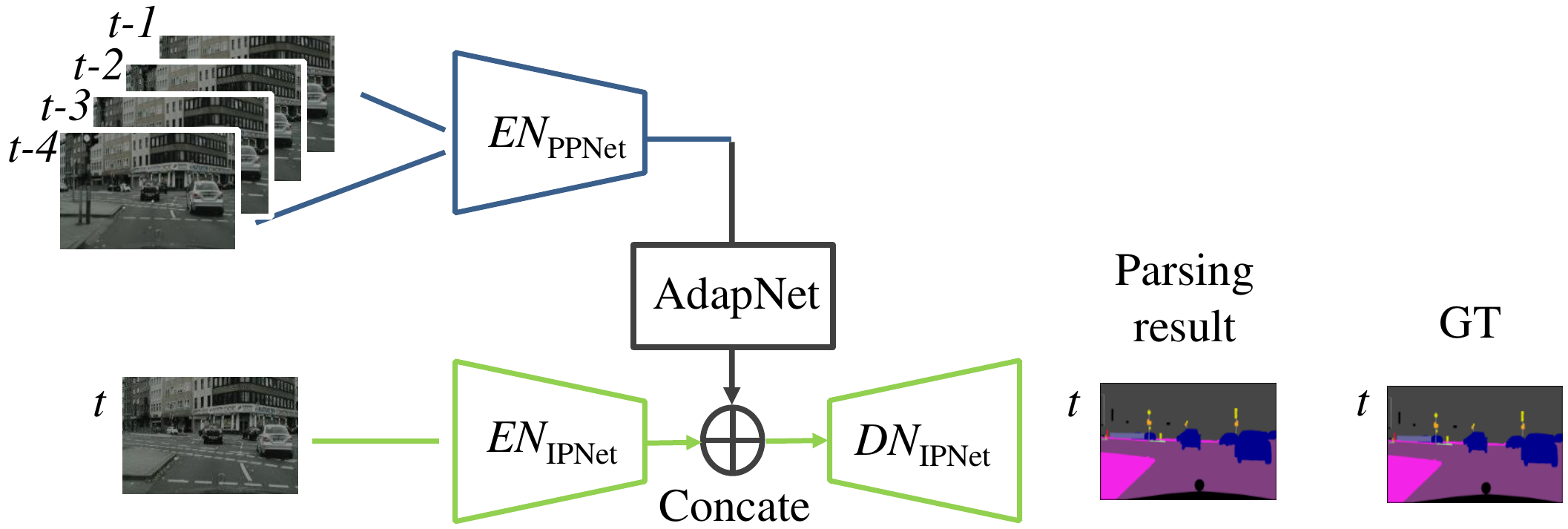}}\\
  \caption{(a) The framework of predictive feature learning. The terms
    \textit{EN} and \textit{DN} represent encoder/decoder network
    respectively. First, FPNet (highlighted in red) learns to \emph{predict
      frame} $t$ given only frames $t\text{-}4$ to $t\text{-}1$ via its
    generator G and discriminator D. Second, PPNet (highlighted in blue)
    performs \emph{predictive parsing} on frame $t$ without seeing it, based on
    its \textit{EN}$_\text{PPNet}$ which is initialized by
    \textit{EN}$_\text{FPNet}$ and connected to \textit{DN}$_\text{PPNet}$. (b)
    The architecture of prediction steering parsing network (PSPNet). Given a
    single input frame $t$, the image parsing network (IPNet) (highlighted in
    green) parses it by integrating the learned features from
    \textit{EN}$_\text{PPNet}$ plus a shallow AdapNet. PPNet and IPNet are
    jointly trained. (c) An important variant of PEARL to verify effectiveness
    of features learned by predictive feature learning. Only the
    \textit{EN}$_\text{FPNet}$ or \textit{EN}$_\text{PPNet}$ is concatenated
    with \textit{EN}$_\text{IPNet}$ through AdapNet. The weights of
    \textit{EN}$_\text{FPNet}$/\textit{EN}$_\text{PPNet}$ are fixed during
    training. Best viewed in color.}
  \label{fig:framework}
  \vspace{-1.5mm}
\end{figure}

Our approach solves these two problems through a novel predictive feature
learning strategy.  We consider the partially-labeled video collection used for
predictive feature learning, denoted as $\{\mathcal{X},\mathcal{Y}\}$, where
$\mathcal{X} = \{ X_1 , \ldots, X_N \}$ denotes the raw video frames and
$\mathcal{Y}=\{Y_{r_1}, \ldots, Y_{r_M}\}$ denotes the provided dense
annotations for a subset of $\mathcal{X}$. Here $M \ll N$ as collecting
large-scale annotation is not easy.  $Y_{r_j}(p,q) \in \{1,\cdots,C\}$ denotes
the ground truth category at location $(p,q)$ in which $C$ is the number of
semantic categories.  Correspondingly, let
$\hat{\mathcal{X}} = \{ \hat X_i , i=1,\cdots,N \}$ and
$\hat{\mathcal{Y}}= \{\hat Y_i , i=1,\cdots,N \}$ denote predicted frames and
predicted parsing maps, respectively. We use $P_i^s=\{X_{i-k}\}_{k=1}^s$ to
denote the $s$ preceding frames ahead of $X_i$. For the first several frames in
a video, we define their preceding set as $X_{i-k}=X_1$ if $i \leq k$.

Video scene parsing can be formulated as seeking a parsing function
$\mathcal{F}$ that maps any frame sequence to the parsing map of the most recent
frame:
{ \setlength{\abovedisplayskip}{10pt} \setlength{\belowdisplayskip}{0pt}
\begin{equation}
\label{eq:define_vsp}
\hat Y_i  = \mathcal F(X_i ,P_i^s ).
\end{equation}
}
\\The above definition reveals the difference between static scene image
parsing and video scene parsing~---~the video scene parsing model $\mathcal F$
has access to historical/temporal information for parsing the current target. We
also want to highlight an important difference between our problem setting and
some existing works \cite{multi-class,budgeted_crf}: instead of using the whole
video (including both past and future frames \textit{w.r.t.}\ $X_i$) to parse
the frame $X_i$, we aim to perform parsing based on causal inference (or online
inference) where only past frames are observable. This setting aligns better
with real-time applications like autonomous driving where future frames cannot
be seen in advance.

\subsection{Predictive Feature Learning}
\label{sec:pfl}
Predictive feature learning aims to learn spatiotemporal features capturing high-level context like object motions and structures from two
consecutive predictive learning tasks, \textit{i.e.}, the frame prediction
and the predictive parsing. Figure~\ref{fig:framework} gives an overview. In the first task, we train an FPNet for
future frame prediction given several past frames by  a new generative
adversarial learning framework developed upon GANs~\cite{goodfellowgan,mathieu}. Utilizing a large amount of unlabeled video sequence data, the
FPNet is capable of learning rich spatiotemporal features to model the
variability of content and dynamics of videos.  Then we further
augment FPNet
through Predictive Parsing, \textit{i.e.}
predicting parsing results of the future frame given previous frames.
This adapts FPNet to another model (called PPNet) suitable for parsing.

\paragraph{Frame Prediction} The architecture of FPNet is illustrated
in Figure \ref{fig:framework_PFL}. It consists of two components,
\textit{i.e.}, the generator (denoted as $G$) which generates future
frame $\hat X_i = G(P_i^s) $ based on its preceding frames $P_i^s$,
and the discriminator (denoted as $D$) which plays against $G$ by
trying to identify predicted frame $\hat X_i $ and the real one
$X_i$. There is an \textit{Encoder}$_{\text{FPNet}}$
(\textit{EN}$_{\text{FPNet}}$ in Figure \ref{fig:framework_PFL}) which
maps the input video sequence to spatiotemporal features and a followed
output layer to produce the RGB values of the predicted frame using the
learned features. Note that \textit{Encoder}$_{\text{FPNet}}$ can
choose any deep networks with various architectures, \textit{e.g.}
VGG16 and Res101. We adapt them to be compatible with video inputs by
using group convolution \cite{alexnet} for the first convolutional
layer, where the group number is equal to the number of input past
frames.

FPNet alternatively trains $D$ and $G$ for predicting frames with
progressively improved quality. Denote learnable parameters of $D$ and
$G$ as $W_D$ and $W_G$ respectively. The objective for training $D$
is to minimize the following binary cross-entropy loss where $G$ is
fixed:
{
 \setlength{\abovedisplayskip}{9pt}
 \setlength{\belowdisplayskip}{0pt}
\begin{equation}
 \label{eq:trainD}
\mathop {\min }\limits_{W_D }\
\ell _D \triangleq - \log (1-D(G(P_i^s ))) - \log D(X_i ).
\end{equation}
} 
\\Minimizing the above loss gives $D$ a stronger discriminative ability to
distinguish the predicted frames $G(P_i^s )$ from real ones $X_i$, enforcing $G$
to predict future frames with higher quality. Towards this target, $G$ learns to
predict future frames more like real ones through {
 \setlength{\abovedisplayskip}{12pt}
 \setlength{\belowdisplayskip}{0pt}
\begin{equation}
\label{eq:trainG}
\mathop {\min }\limits_{W_G }\  \ell _G = \ell _{rec} + \lambda_{adv}\ell _{adv},
\end{equation}
}
\\where
{
  \setlength{\abovedisplayskip}{8pt} \setlength{\belowdisplayskip}{0pt}
\begin{equation*}
\label{eq:ladv}
\ell_{rec} = \| {X_j  - \hat X_j } \|_2, \text{ and }
\ell_{adv} = - \log D(G(\hat P_i^s )).
\end{equation*}
}\\Minimizing the combination of reconstruction loss and adversarial loss
supervises $G$ to predict the frame that looks both similar to its corresponding
real frame and sufficiently authentic to fool the strong competitor $D$.  Our
proposed frame prediction model is substantially different from vanilla GAN and
more tailored for VSP problems. The key difference lies in the generator $G$
that takes past frame sequence $P_i^s$ as input to predict the future frame,
instead of crafting new samples completely from random noise as vanilla
GANs. Therefore, the future frames coming from such ``temporally conditioned''
FPNet should present certain temporal consistency with past frames. On the other
hand, FPNet can learn representations containing implicit temporal cues desired
for solving VSP problems.

\begin{figure*}[t!]
	\centering     
	\includegraphics[width=\linewidth]{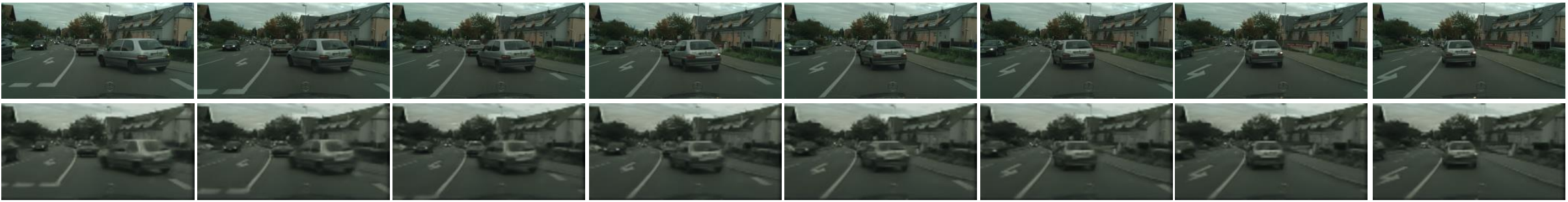}
	\caption{Example frame predictions of FPNet on Cityscape val
          set. \textbf{Top}: ground truth video sequence. \textbf{Bottom}: frame
          prediction of FPNet. FPNet produces visually similar frames with the
          ground truth, demonstrating that FPNet learns robust spatiotemporal
          features to model the structures of objects (\textit{building}) and
          stuff (\textit{vegetation}), and motion information of moving objects
          (\textit{cars}), both of which are critical for VSP problems.}
	\label{fig:fp_vis}
\end{figure*}

As illustrated by Figure \ref{fig:fp_vis}, FPNet produces real-looking
frame predictions by learning both the content and dynamics in
videos. By comparing with the ground truth frame, the prediction frame
resembles both the structures of objects/stuff like building/vegetation
and the motion trajectories of cars, demonstrating that FPNet learns
robust and generalized spatiotemporal features from video data.

In our experiments, we use a GoogLeNet~\cite{googlenet} as $D$ and we try both
Res101 and VGGNet for $G$. More details are given in Section \ref{sec:settings}.

\paragraph{Predictive Parsing}
The features learned by FPNet so far are trained for video frame generation.  To
adapt these spatiotemporal features to VSP problems, FPNet performs the second
predictive learning task, \textit{i.e.}, predicting the parsing map of one frame
$X_i$ given only its preceding frames $P_i^s$ (without seeing the frame to
parse). This predictive parsing task is very challenging as we do not have any
information of the current video frame.

Also, directly training FPNet for this predictive parsing task from scratch will
not succeed. There are no enough data with annotations for training a good model
free of over-fitting. Thus, training FPNet for frame prediction at first gives a
good starting model for accomplishing the second task. In this perspective,
frame prediction training is important for both spatiotemporal feature learning
and feature adaption.

Details on how FPNet performs the predictive parsing task are given in
Figure~\ref{fig:framework_PFL}. For predicting parsing maps, we modify the
architecture of FPNet as follows. We remove the $D$ component from FPNet as well
as the output layer in $G$. Then we add a deconvNet (\textit{i.e.}
\textit{DN}$_{\text{PPNet}}$ in the figure) on top of modified $G$, which
produces the parsing map sharing the same size with frames. We call this new
FPNet as PPNet, short for Predictive Parsing Net.

More details about the structure of PPNet are given in
Section~\ref{sec:settings}.  Based on the notations given in
Section~\ref{sec:notations}, the objective function for training PPNet is
defined as
{
 \setlength{\abovedisplayskip}{8pt}
 \setlength{\belowdisplayskip}{0pt}
\begin{equation}
  \label{eq:trainPPNet}
  \ell_{\text{PPNet}}  =  - \sum\limits_{(p,q) \in X_{r_j } } {\omega_{Y_{r_j } (p,q)} h_{Y_{r_j } (p,q)} (W_{\text{PPNet}},P_{r_j }^s )},
\end{equation}
}
\\where $h_{Y_{r_j } (p,q)}$ denotes the per-pixel logarithmic probability
predicted by PPNet for the category label $Y_{r_j}(p,q)$. We introduce the
weight vector $\omega$ to balance scene classes with different frequency in the
training set. We will further discuss its role in the experiments.

Examples of predicted parsing maps from PPNet are shown in Figure
\ref{fig:vis_cmp} (the third row). Compared with parsing results from
a traditional CNN parsing model on the single frame, the parsing maps
of PPNet present two distinct properties. First, the parsing maps are
temporally smooth which are reflected in the temporally consistent
parsing results of regions like building where CNN models produce
noisy and inconsistent parsing results. This demonstrates the PPNet
indeed learns the temporally continuous dynamics from the video
data. Secondly, PPNet tends to miss objects of small sizes,
\textit{e.g.}, transport signs and poles. One reason is the inevitable
blurry prediction \cite{mathieu} since the high
frequency spectrum is prone to being smoothed. This problem can be
mitigated by parsing at multiple scales \cite{deeplabv2} and we will
investigate in the future.

In contrast, the conventional image parsing network relies on locally
discriminative features such that it can capture small objects. However, due to
lacking temporal information, its produced parsing maps　 are noisy and lack
temporal consistency with past frames. Above observations motivate us to combine
the strengths of PPNet and the CNN-based image parsing model to improve the
overall VSP performance. Therefore, we develop the following prediction steering
parsing architecture.

\subsection{Prediction Steering Parsing}
\label{sec:psp}
To take advantage of the temporally consistent spatiotemporal features learned
by PPNet, we propose a novel architecture to integrate PPNet and the traditional
image parsing network (short for IPNet) into a unified framework, called PSPNet,
short for Prediction Steering Parsing Net.

As illustrated in Figure \ref{fig:framework_PSP}, PSPNet has two inter-connected
branches: one is the PPNet for predictive feature learning and the other is
IPNet for frame-by-frame parsing.  Similar to FPNet, the IPNet can also be
chosen freely from any existed image parsing networks, \textit{e.g.}  FCN
\cite{fullyconvseg} and DeepLab \cite{deeplabv2}. As a high-level description,
IPNet consists of two components, a feature encoder
\textit{Encoder}$_\text{IPNet}$ (\textit{EN}$_{\text{IPNet}}$) which transforms
the input frame to dense pixel features and a deconvNet
(\textit{DN}$_{\text{IPNet}}$) that produces per-pixel parsing map. Through an
AdapNet (a shallow CNN), PPNet communicates its features to IPNet and steers the
overall parsing process. In this way, the integrated features within IPNet gain
two complementary properties, \textit{i.e.}, descriptiveness for the temporal
context and discriminability for different pixels within a single
frame. Therefore the overall PSPNet model is capable of producing more accurate
video scene parsing results than both PPNet and IPNet. Formally, the objective
function of training PSPNet end-to-end is defined as {
  \setlength{\abovedisplayskip}{8pt} \setlength{\belowdisplayskip}{0pt}
\begin{align}
\label{eq:trainPSPNet}
  L_{\text{PSPNet}} = - \sum\limits_{(p,q) \in X_{r_j } } &\omega_{Y_{r_j } (p,q)}
                                                            \left( h_{Y_{r_j } (p,q)} (W_{\text{PPNet}},P_{r_j }^s )\right. \nonumber \\
                                                          & \left.+ \lambda _{\text{IP}}
                                                            f_{Y_{r_j } (p,q)} (W_{\text{IPNet}},X_{r_j } )\right),
\end{align}
}
\\where $f_{Y_{r_j } (p,q)} (W_{\text{IPNet}},X_{r_j } )$ denotes the
per-pixel logarithmic probability produced by the \textit{DN}$_{\text{IPNet}}$
and $\lambda_{\text{IP}}$ balances the effect of PPNet and IPNet. We start
training PSPNet by taking the trained PPNet in predictive parsing as
initialization. We find this benefits the convergence of PSPNet training. In
Section~\ref{sec:ablation}, we give more empirical studies.

Now we proceed to explain the role of AdapNet. There are two
disadvantages by naively combining the intermediate features of PPNet
and IPNet. Firstly, since the output features from those two feature
encoders generally have different distributions, naively concatenating
features harms the final performance as the ``large'' features
dominate the ``smaller'' ones. Although during training, the
followed weights in deconvNet may adjust accordingly, it requires
careful tuning of parameters thus is subject to trial and error to find
the best settings. Similar observations have been made in previous
literature~\cite{parsetnet}. However, different from~\cite{parsetnet}
which uses a normalization layer to tackle the scale problem, we use a
more powerful AdapNet to transform the features of PPNet to be with
proper norm and scale. Secondly, the intermediate features of PPNet
and IPNet are with different semantic meanings, which means they reside
in different manifold spaces. Therefore naively combining them
increases the difficulty of learning the transformation from feature
space to parsing map in the followed \textit{DN}$_{\text{IPNet}}$. By
adding the AdapNet to convert the feature space in advance, it eases the
training of \textit{DN}$_{\text{IPNet}}$. Detailed explorations of the
architecture of AdapNet follows in Section~\ref{sec:ablation}.

\subsection{Discussion}
\paragraph{Unsupervised Feature Learning}
Currently we train FPNet in a pseudo semi-supervised way, \textit{i.e.}
initialize $G$ and $D$ with ImageNet pre-trained models for faster
training. Without using the pre-trained models, our approach becomes an
unsupervised feature learning one. We also investigate the fully unsupervised
learning strategy of FPNet in the experiments.  The resulting FPNet is denoted
as FPNet$_{\text{VGG11}}^*$. As shown in Table \ref{table:step123_cmp},
FPNet$_{\text{VGG11}}^*$ performs similarly well as FPNet$_{\text{VGG16}}$ using
the pre-trained VGG16 model. In the future, we will perform unsupervised
learning on FPNet using deeper architectures and further improve its ability.

\paragraph{Proactive Parsing}
Our predictive learning approach recalls another challenging but attractive
task, \textit{i.e.}, to predict the future parsing maps for a few seconds
without seeing them, only based on past frames. Achieving this allows autonomous
vehicles or other parsing-demanded devices to receive parsing information ahead
and get increased buffer time for decision making.  Our approach indeed has the
potential to accomplish this proactive parsing task.  As one can observe from
Figure~\ref{fig:vis_cmp}, the predicted parsing maps capture the temporal
information across frames, such as motions of cars, and the predicted parsing
map reflects such dynamics and shows roughly correct prediction.  In the future,
we will investigate how to enhance the performance of our approach on predictive
parsing to get higher-quality and longer-term future results.

\section{Experiments}
\label{sec:experiments}
\subsection{Settings and Implementation Details}
\label{sec:settings}
\paragraph{Datasets} Since PEARL tackles the scene parsing problem with temporal
context, we choose Cityscapes \cite{cityscape} and Camvid \cite{camvid} for
evaluation. Both datasets provide annotated frames as well as adjacent frames,
suitable for testing the temporal modeling ability. Cityscapes is a large-scale
dataset containing finely pixel-wise annotations on 2,975/500/1,525
train/val/test frames with 19 semantic classes and another 20,000 coarsely
annotated frames. Each finely annotated frame is sampled from the 20th frame of
a 30-frame video clip in the dataset, giving in total 180K frames. Since there
are no video data provided for the coarsely annotated frames, we only use finely
annotated ones for training PEARL. Every frame in Cityscapes has a resolution of
1024$\times$2048 pixels.

The Camvid dataset contains 701 color images with annotations on 11 semantic
classes. These images are extracted from driving videos captured at daytime and
dusk. Each video contains 5,000 frames on average, with a resolution of
720$\times$960 pixels, giving in total 40K frames.

\paragraph{Baselines}
We conduct experiments to compare PEARL with two baselines which use different
deep network architectures.
\begin{itemize}[leftmargin=*]
	\setlength\itemsep{0em}
      \item{\textbf{VGG16-baseline}}\quad Our VGG16-baseline is built upon
        DeepLab~\cite{chen2014semantic}. We make the following modifications. We
        add three deconvolutional layers after \texttt{fc7} to learn better
        transformations to label maps. The architectures of three added
        deconvolutional layers in VGG16-baseline are
        $O(256) \mhyphen K(4) \mhyphen S(2)\mhyphen P(1)$,
        $O(128) \mhyphen K(4) \mhyphen S(2)\mhyphen P(1)$ and
        $O(64) \mhyphen K(4) \mhyphen S(2)\mhyphen P(1)$ respectively. Here
        $O(n)$ denotes $n$ output feature maps, $K(n)$ denotes the kernel size
        of $n \times n$, $S(n)$ denotes a stride of length $n$ and $P(n)$
        denotes the padding size of $n$. The layers before \texttt{fc7}
        (included) constitute the encoder network (\textit{EN} in
        Figure~\ref{fig:framework}) and the other layers form the decoder
        network (\textit{DN} in Figure~\ref{fig:framework}).
      \item{\textbf{Res101-baseline}}\quad Our Res101-baseline is modified from
        \cite{residual} by adapting it to a fully convolutional network
        following \cite{fullyconvseg}. Specifically, we replace the average
        pooling layer and the 1000-way classification layer with a fully
        convolutional layer to produce dense label maps. Also, we modify
        \texttt{conv5\textunderscore1}, \texttt{conv5\textunderscore2} and
        \texttt{conv5\textunderscore3} to be dilated convolutional layers by
        setting their dilation size to be 2 to enlarge the receptive fields. As
        a result, the output feature maps of \texttt{conv5\textunderscore3} have
        a stride of 16. Following \cite{fullyconvseg}, we utilize high-frequency
        features learned in bottom layers by adding skip connections from
        \texttt{conv1}, \texttt{pool1}, \texttt{conv3\textunderscore 3} to
        corresponding up-sampling layers to produce label maps with the same
        size as input frames. The layers from \texttt{conv1} to
        \texttt{conv5\textunderscore3} belong to \textit{EN} while the other
        layers belong to \textit{DN}. Following~\cite{highperformance}, we also
        use hard training sample mining to reduce over-fitting.
\end{itemize}

Note that IPNets in PEARL share the same network architectures as baseline
models. The encoder networks in FPNet/PPNet and the decoder network in PPNet
share the same network architectures of the encoder network and the decoder
network in baseline models, respectively.

\paragraph{Evaluation Metrics} Following previous practice, we use the mean IoU
(mIoU) for Cityscapes, and per-pixel accuracy (PA) and average per-class
accuracy (CA) for Camvid. In particular, mIoU is defined as the pixel
intersection-over-union (IOU) averaged across all categories; PA is defined as
the percentage of all correctly classified pixels; and CA is the average of all
category-wise accuracies.

\paragraph{Implementation Details}
Throughout the experiments, we set the number of preceding frames of each frame
as 4, \emph{i.e.}, $s=4$ in $P_i^s$ (ref. Section~\ref{sec:notations} in the
main text). When training FPNet, we randomly select frame sequences from those
with a length of 4 and also enough movement (the $\ell_2$ distance between the
raw frames is larger than a threshold $230$). In this way, we finally obtain
35K/8.8K sequences from Cityscapes and Camvid respectively. The input frames for
training FPNet are all normalized such that values of their pixels lie between
-1 and 1. For training PPNet and PSPNet, we only perform mean value subtraction
on the frames. For training PPNet, we select the 4 frames before the annotated
images to form the training sequences, where the frames are required to have
sufficient motion, consistent with FPNet.

We perform random cropping and horizontal flipping to augment the training
frames for FPNet, PPNet and PSPNet. In addition, for training FPNet, the
temporal order of a sequence (including the to-predict frame and 4 preceding
frames) is randomly reversed to model various dynamics in videos. The
hyperparameters in PEARL are fixed as $\lambda_{adv}=0.2$ in
Eq. \eqref{eq:trainG} and $\lambda_{\text{IP}}=0.3$ in
Eq. \eqref{eq:trainPSPNet} and the probability threshold of hard training sample
mining in Res101-baseline as 0.9. The values are chosen through
cross-validation.

Since the class distribution is extremely unbalanced, we increase the weight of
rare classes during training, similar to~\cite{farabet2013learning,mpf,dag}. In
particular, we adopt the re-weighting strategy in~\cite{dag}. The weight for the
class $\mathbf{y}_{i,j}$ is set as
$\omega _{\mathbf{y}_{i,j} } = 2^{\left\lceil {\log 10(\eta /f_{\mathbf{y}_{i,j}
      } )} \right\rceil }$
where $f_{\mathbf{y}_{i,j}}$ is the frequency of class $\mathbf{y}_{i,j}$ and
$\eta$ is a dataset-dependent scalar, which is defined according to the
85\%/15\% frequent/rare classes rule.

All of our experiments are carried out on NVIDIA Titan X and Tesla M40 GPUs
using Caffe library.　\vspace{-6.1mm}

\paragraph{Computational Efficiency} Since no post-processing is required in
PEARL, the running time of PEARL$_\text{Res101}$ for obtaining the parsing map
of a frame with resolution 1,024$\times$2,048 is only 0.8 seconds on a modern
GPU, among the fastest methods in existing works.

\begin{figure*}[t!]
 \captionsetup[subfigure]{labelformat=empty}
  \centering
  \subfloat[]{
    \includegraphics[width=\linewidth]{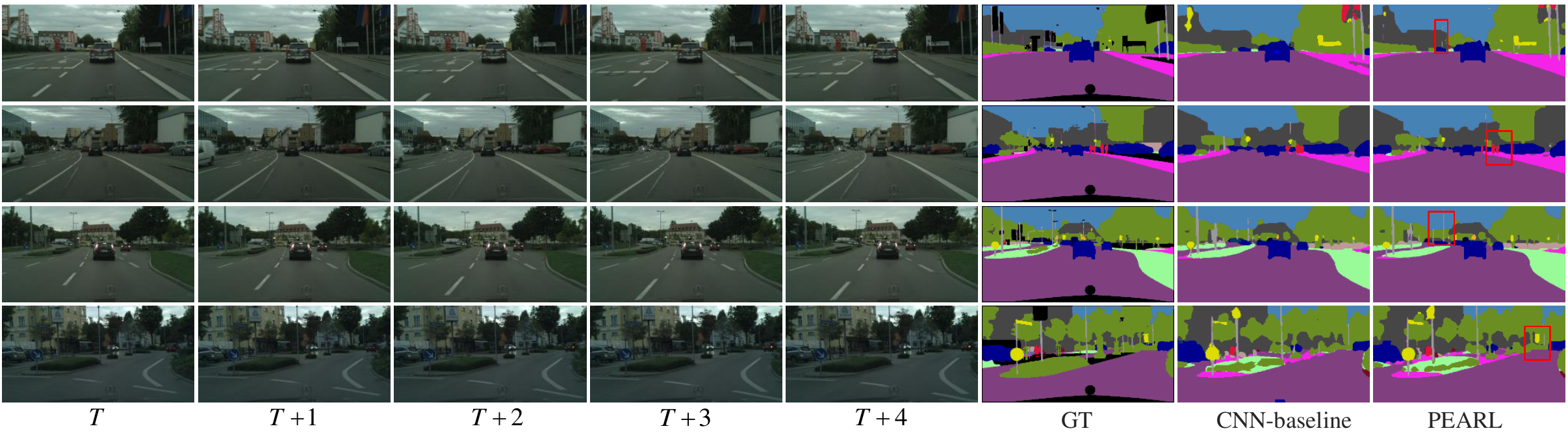}}
  \phantomcaption\vspace{-6mm}\\
  \caption{Examples of parsing results of PEARL on Cityscape val set. The first
    five images in each row represents a different video sequence, which are
    followed by ground truth annotations, the parsing map of the baseline model
    and the parsing map of our proposed PEARL, all for frame $T$+4. It is
    observable that PEARL produces more smooth label maps and shows stronger
    discriminability for small objects (highlighted in red boxes) compared to
    the baseline model. Best viewed in color and zoomed pdf.}
	\label{fig:qualitative_vis}
\end{figure*}

\subsection{Results}

Examples of parsing maps produced by PEARL are illustrated in
Figure~\ref{fig:qualitative_vis}, where VGG16 is used in both the baseline model
and PEARL. As can be seen from Figure~\ref{fig:qualitative_vis}, compared with
the baseline model, PEARL produces smoother parsing maps, \textit{e.g.} for the
class \textit{vegetation}, and stronger discriminability for small objects,
\text{e.g.} for the classes \textit{pole}, \textit{pedestrian}, \textit{traffic
  sign} as highlighted in red boxes. Such improvements are attributed to the
capability of PEARL to learn the temporally consistent features and
discriminative features for local pixel variance simultaneously.

\subsubsection{Cityscapes}
\label{sec:ablation}

\paragraph{Ablation Analysis} We investigate contribution of each component of
our approach.

\textit{(1) Predictive Feature Learning}.\quad To investigate the contribution
of the two predictive feature learning networks, FPNet and PPNet, we conduct
three experiments. The comparison results are listed in
Table~\ref{table:step123_cmp}, where the front-CNN of FPNet and PPNet both use
the VGG16 architecture.

First, we would like to verify the effectiveness of the features learned by
FPNet for VSP. We concatenate the output features (denoted as
\verb'feat'$_{\text{FPNet}}$) of \textit{Encoder}$_{\text{FPNet}}$ with the
output features of \textit{Encoder}$_{\text{IPNet}}$ in IPNet, as shown in
Figure~\ref{fig:framework_variant}, and fix the
\textit{Encoder}$_{\text{FPNet}}$ during training. In this way, the FPNet only
extracts spatiotemporal features for the IPNet. As can be seen from
Table~\ref{table:step123_cmp}, by combining \verb'feat'$_{\text{FPNet}}$, the
mIoU increases from 63.4 (of the VGG16-baseline) to 68.6, demonstrating FPNet
indeed learns useful spatiotemporal features through frame prediction.

Similarly, we replace \textit{Encoder}$_{\text{FPNet}}$ in the above experiment
with \textit{Encoder}$_{\text{PPNet}}$ to investigate the influence of PPNet on
final performance. In the experiment, the per-pixel cross-entropy loss layer of
PPNet is removed and the weights of \textit{Encoder}$_{\text{PPNet}}$ are
fixed. As illustrated in Table~\ref{table:step123_cmp}, combining IPNet with
\verb'feat'$_{\text{PPNet}}$ further increases the mIoU by 0.5 compared to
combining \verb'feat'$_{\text{FPNet}}$, demonstrating features learned by PPNet
from predictive parsing are useful for VSP.

Finally, we look into the effectiveness of joint training of PPNet and IPNet. It
is observed from Table~\ref{table:step123_cmp} that the resulting model,
\textit{i.e.} PEARL$_\text{VGG16}$ achieves the best performance, benefiting
from the joint end-to-end training strategy.

\begin{table}
  \cprotect\caption{Comparative study of effects of FPNet and PPNet on final
    performance of PEARL over Cityscapes val set. The front models of FPNet and PPNet are VGG16 for fair comparison. \verb'feat'$_{\text{Net}}$ means only the output features of ``Net'' are combined with IPNet. $\text{FPNet}^*$ means FPNet is trained from random initialization.}
	\centering
	\begin{tabular}{lc} \hline Methods       & {mIoU}   \\
          \hline
          VGG16-baseline                                & 63.4            \\
          \hline
          \verb'feat'$_{\text{FPNet}_\text{VGG16}}$ + IPNet$_\text{VGG16}$        & 68.6   \\
          \verb'feat'$_{\text{FPNet}_{\text{VGG11}}^*}$ + IPNet$_\text{VGG16}$    & 68.4   \\
          \verb'feat'$_{\text{PPNet}_\text{VGG16}}$ + IPNet$_\text{VGG16}$        & 69.1   \\
          PEARL$_\text{VGG16}$                                                    &\textbf{69.8} \\
          \hline
          \label{table:step123_cmp}
	\end{tabular}
\end{table}

\begin{table}
  \cprotect\caption{Comparative study of PEARL with optical flow based method on two deep networks: VGG16 and Res101 to verfity the superiority of PEARL on modeling temporal information. \verb'feat'$_{\text{OF}}$ means the optical flow maps calculated by epic flow \cite{epicflow}.}
	\label{table:op_cmp}
	\small
	\centering
	\begin{tabular}{lc} \hline Methods       & {mIoU}   \\
          \hline
          VGG16-baseline                                        & 63.4            \\
          \verb'feat'$_\text{OF}$ + IPNet$_\text{VGG16}$        & 64.5            \\
          PEARL$_\text{VGG16}$		                            & \textbf{69.8}   \\
          \hline
          \hline
          Res101-baseline                                       & 72.5            \\
          \verb'feat'$_\text{OF}$ + IPNet$_\text{Res101}$       & 72.7            \\
          PEARL$_\text{Res101}$	                                & \textbf{74.9}   \\
          \hline
	\end{tabular}
\end{table}

\begin{table*}
	\centering
	\caption{Performance comparison of PEARL with state-of-the-arts on Cityscapes \emph{test} set. Note for fast inference, single scale testing is used in PEARL without any post-processing like CRF.}
	\label{table:cityscapes_test}
	\small
	\setlength{\tabcolsep}{1.7pt}
	\begin{tabular}{l|ccccccccccccccccccc|c}
          \hline
          \multicolumn{1}{c|}{Methods}   & \multicolumn{1}{c}{\rotatebox[origin=l]{90}{road}} & \multicolumn{1}{c}{\rotatebox[origin=l]{90}{sidewalk}} & \multicolumn{1}{c}{\rotatebox[origin=l]{90}{building}} & \multicolumn{1}{c}{\rotatebox[origin=l]{90}{wall}} & \multicolumn{1}{c}{\rotatebox[origin=l]{90}{fence}} & \multicolumn{1}{c}{\rotatebox[origin=l]{90}{pole}} & \multicolumn{1}{c}{\rotatebox[origin=l]{90}{traffic light}} & \multicolumn{1}{c}{\rotatebox[origin=l]{90}{traffic sign}} & \multicolumn{1}{c}{\rotatebox[origin=l]{90}{vegetation}} & \multicolumn{1}{c}{\rotatebox[origin=l]{90}{terrain}} & \multicolumn{1}{c}{\rotatebox[origin=l]{90}{sky}} & \multicolumn{1}{c}{\rotatebox[origin=l]{90}{person}} & \multicolumn{1}{c}{\rotatebox[origin=l]{90}{rider}} & \multicolumn{1}{c}{\rotatebox[origin=l]{90}{car}} & \multicolumn{1}{c}{\rotatebox[origin=l]{90}{truck}} & \multicolumn{1}{c}{\rotatebox[origin=l]{90}{bus}} & \multicolumn{1}{c}{\rotatebox[origin=l]{90}{train}} & \multicolumn{1}{c}{\rotatebox[origin=l]{90}{motorcycle}} & \multicolumn{1}{c|}{\rotatebox[origin=l]{90}{bicycle}} & \multicolumn{1}{c}{mIoU}  \\ \hline 

          \multirow{1}{*}{FCN\textunderscore8s~\cite{fullyconvseg}} & 97.4 & 78.4 & 89.2 & 34.9 & 44.2 & 47.4 & 60.1 & 65.0 & 91.4 & 69.3 & 93.9 & 77.1 & 51.4 & 92.6 & 35.3 & 48.6 & 46.5 & 51.6 & 66.8 & 65.3 \\ 

          \multirow{1}{*}{DPN~\cite{liu2015semantic}} & 97.5 & 78.5 & 89.5 & 40.4 & 45.9 & 51.1 & 56.8 & 65.3 & 91.5 & 69.4 & 94.5 & 77.5 & 54.2 & 92.5 & 44.5 & 53.4 & 49.9 & 52.1 & 64.8 & 66.8 \\

          \multirow{1}{*}{Dilation10~\cite{yu2015multi}} & 97.6 & 79.2 & 89.9 & 37.3 & 47.6 & 53.2 & 58.6 & 65.2 & 91.8 & 69.4 & 93.7 & 78.9 & 55.0 & 93.3 & 45.5 & 53.4 & 47.7 & 52.2 & 66.0 & 67.1 \\

          \multirow{1}{*}{DeepLab~\cite{deeplabv2}} & 97.9 & 81.3 & 90.3 & 48.8 & 47.4 & 49.6 & 57.9 & 67.3 & 91.9 & 69.4 & 94.2 & 79.8 & 59.8 & 93.7 & 56.5 & 67.5 & 57.5 & 57.7 & 68.8 & 70.4   \\ 

          \multirow{1}{*}{Adelaide~\cite{adelaide16}} & 98.0 & 82.6 & 90.6 & 44.0 & 50.7 & 51.1 & 65.0 & 71.7 & 92.0 & \textbf{72.0} & 94.1 & 81.5 & 61.1 & 94.3 & 61.1 & 65.1 & 53.8 & 61.6 & 70.6 & 71.6  \\ 

          \multirow{1}{*}{LRR-4X~\cite{lrr4x}} & 97.9 & 81.5 & 91.4 & \textbf{50.5} & 52.7 & 59.4 & \textbf{66.8} & \textbf{72.7} & 92.5 & 70.1 & 95.0 & 81.3 & 60.1 & 94.3 & 51.2 & 67.7 & 54.6 & 55.6 & 69.6 & 71.8 \\

          \multirow{1}{*}PEARL\footnotemark (ours) & \textbf{98.3} & \textbf{83.9} & \textbf{91.6} & 47.6 & \textbf{53.4} & \textbf{59.5} & \textbf{66.8} & 72.5 & \textbf{92.7} & 70.9 & \textbf{95.2} & \textbf{82.4} & \textbf{63.5} & \textbf{94.7} & \textbf{57.4} & \textbf{68.8} & \textbf{62.2} & \textbf{62.6} & \textbf{71.5} & \textbf{73.4} \\ 
          \hline
	\end{tabular}
\end{table*}

\begin{table}[t]
  \cprotect\caption{Comparison with state-of-the-arts on Cityscapes val set. Single scale testing is used in PEARL w/o post-processing as CRF.}
	\label{table:val_cmp}
	\small
	\centering
	\begin{tabular}{lc} \hline Methods       & mIoU   \\
          \hline
          VGG16-baseline (ours)                               & 63.4            \\
          FCN~\cite{fullyconvseg}                             & 61.7            \\
          Pixel-level Encoding~\cite{uhrig2016pixel}          & 64.3            \\
          DPN~\cite{liu2015semantic}                          & 66.8            \\
          Dilation10~\cite{yu2015multi}                       & 67.1            \\
          DeepLab-VGG16~\cite{deeplabv2}                      & 62.9            \\
          Deep Structure~\cite{adelaide16}                    & 68.6            \\
          Clockwork FCN~\cite{shelhamer2016clockwork}         & 64.4            \\
          PEARL$_\text{VGG16}$ (ours)		                  & \textbf{69.8}   \\
          \hline
          \hline
          Res101-baseline (ours)                              & 72.5            \\
          DeepLab-Res101~\cite{deeplabv2}                     & 71.4            \\
          PEARL$_\text{Res101}$ (ours)		                  & \textbf{74.9}   \\
          \hline
	\end{tabular}
\end{table}

\textit{(2) Comparison with Optical Flow Methods}.\quad To verify the
superiority of PEARL on learning the temporal information specific for VSP, we
compare PEARL with other temporal context modeling methods. First, we naively
pass each frame in $P_i^s$ and $X_i$ through baseline models (both
VGG16-baseline and Res101-baseline) and merge their probability maps to obtain
the final label map of $X_i$. It is verified by experiments that such methods
achieve worse performance than baseline models due to their weakness of
utilizing
\footnotetext{\scriptsize{https://www.cityscapes-dataset.com/method-details/?submissionID=328}}
temporal information and the noisy probability map produced for each
frame. Since optical flow is naturally capable of modeling the temporal
information in videos, we use it as a strong baseline to compete with PEARL. We
employ the epic flow~\cite{epicflow} for computing all optical flows. Then we
warp the parsing map of the frame $X_{i-1}$ and merge it with that of the frame
$X_i$, according to the optical flow calculated between these two frames. In
this way, the temporal context is modeled explicitly via optical flow. This
method performs better than the last method but its performance is still
inferior to baseline models. It is because the CNN models produce the parsing
map without knowing the temporal information during training.

Then we conduct the third experiment by concatenating the optical flows
calculated from $X_{i-1}$ to $X_i$ with the frame $X_i$, which forms 5-channel
raw data (RGB plus X/Y channels of optical flow). Based on optical flow
augmented data, we re-train baseline models. During training, each kernel in the
first convolutional layer of baseline models is randomly initialized for the
weights corresponding to the X/Y channels of optical flow. This method is
referred to as ``\verb'feat'$_\text{OF}$ + IPNet''. The comparative results of
``\verb'feat'$_\text{OF}$ + IPNet'' and PEARL using VGG16 and Res101 are
displayed in Table~\ref{table:op_cmp}. From the results, one can observe
``\verb'feat'$_\text{OF}$ + IPNet'' achieves higher performance than baselines
models as it uses temporal context during training. Notably, PEARL significantly
beats ``\verb'feat'$_\text{OF}$ + IPNet'' on both network architectures, proving
its superior ability to model temporal information for VSP problems.

\textit{(3) Ablation Study of AdapNet}\quad As introduced in Section
\ref{sec:psp}, AdapNet improves the performance of PEARL by learning the latent
transformations from \textit{Encoder}$_{\text{PPNet}}$ to
\textit{Encoder}$_{\text{IPNet}}$. In our experiments, the AdapNet contains one
convolutional layer followed by ReLU. The kernel size of the convolutional layer
is 1 and the number of kernels is equal to that of output channels of
\textit{Encoder}$_{\text{PPNet}}$. Compared to the PEARL w/o AdapNet, adding
AdapNet brings 1.1/0.3 mIoU improvements for PEARL$_\text{VGG16}$ and
PEARL$_\text{Res101}$, respectively. We also conduct experiments by increasing
convolutional layers of AdapNet, but only observe marginal improvements. Since
deeper AdapNet brings more computation cost, we use AdapNet with one
convolutional layer.

\paragraph{Comparison with State-of-the-arts}
The comparison of PEARL with other state-of-the-arts on Cityscapes val set is
listed in Table~\ref{table:val_cmp}, from which one can observe PEARL achieves
the best performance among all compared methods on both network
architectures. Note loss re-weighting is not used on this dataset.

Specifically, PEARL$_\text{VGG16}$ and PEARL$_\text{Res101}$ significantly
improve the corresponding baseline models by 6.4/2.4 mIoU,
respectively. Notably, compared with \cite{shelhamer2016clockwork} which
proposed a temporal skip network based on VGG16 for video scene parsing,
PEARL$_\text{VGG16}$ beats it by 5.4 in terms of mIoU. We also note that
different from other methods which extensively modify VGG16 networks to enhance
its discriminative power for image parsing,
\textit{e.g.}~\cite{deeplabv2,adelaide16}, our PEARL$_\text{VGG16}$ is built on
vanilla VGG16 architecture. Thus it is reasonable to expect further improvement
on the VSP performance by using more powerful base network
architectures. Furthermore, we submit PEARL$_\text{Res101}$ to the online
evaluation server of Cityscapes to compare with other state-of-the-arts on
Cityscapes test set. As shown in Table~\ref{table:cityscapes_test}, our method
achieves the best performance among all top methods which have been published
till the time of submission. Note in inference, PEARL only uses single-scale
testing without CRF post-processing for the sake of fast inference.

\subsubsection{Camvid}

We further investigate the effectiveness of PEARL on Camvid. Its result and the
best results ever reported on this dataset are listed in Table
\ref{table:camvid}. Following \cite{mpf,dag}, loss re-weighting is used on this
dataset. One can observe that PEARL performs much better than all competing
methods~---~significantly improving the PA/CA of the baseline model
(Res101-basline) by 1.6\%/2.5\% respectively, once again demonstrating its
strong capability of improving VSP performance. Notably, compared to the optical
flow based methods \cite{RTDF} and \cite{liu2015semantic} which utilize CRF to
model temporal information, PEARL shows great advantages in performance,
verifying its superiority in modeling the spatiotemporal context for VSP
problems.

\begin{table}[h]
  \caption{Comparison with the state-of-the-art methods on CamVid.}
  \label{table:camvid}
   \small
  \centering
  \begin{tabular}{lcc}
    \hline
    Methods       & {PA(\%)}       &{CA(\%)}  \\
    \hline
    Res101-baseline (ours)                                    & 92.6            & 80.0 \\
    Ladicky \emph{et al.}~\cite{ladicky2010and}               & 83.8            & 62.5   \\
    SuperParsing~\cite{tighe2010superparsing}                 & 83.9            & 62.5   \\
    DAG-RNN~\cite{dag}                                        & 91.6            & 78.1   \\
    MPF-RNN~\cite{mpf}                                        & 92.8            & 82.3 \\
    Liu \emph{et al.}~\cite{liu2015semantic}                  & 82.5            & 62.5 \\ RTDF~\cite{RTDF}                                          & 89.9            & 80.5 \\
    \hline
    PEARL (ours)                                              & \bf{94.2}       & \bf{82.5} \\
    \hline
    \end{tabular}
\end{table}

\section{Conclusion}
We proposed a predictive feature learning method for effective video scene
parsing. It contains two novel components: predictive feature learning and
prediction steering parsing. The first component learns spatiotemporal features
by predicting future frames and their parsing maps without requiring extra
annotations. The prediction steering parsing architecture then guides the single
frame parsing network to produce temporally smooth and structure preserving
results by using the predictive feature learning outputs. Extensive experiments
on Cityscapes and Camvid fully demonstrated the effectiveness of our approach.
\newpage

\bibliographystyle{ieee}
\bibliography{mybibfile}

\begin{appendices}
\section{Qualitative Evaluation of PEARL}
\subsection{More Results of Frame Prediction from FPNet in PEARL}
\begin{figure*}[t!]
  \centering
	\includegraphics[width=\linewidth]{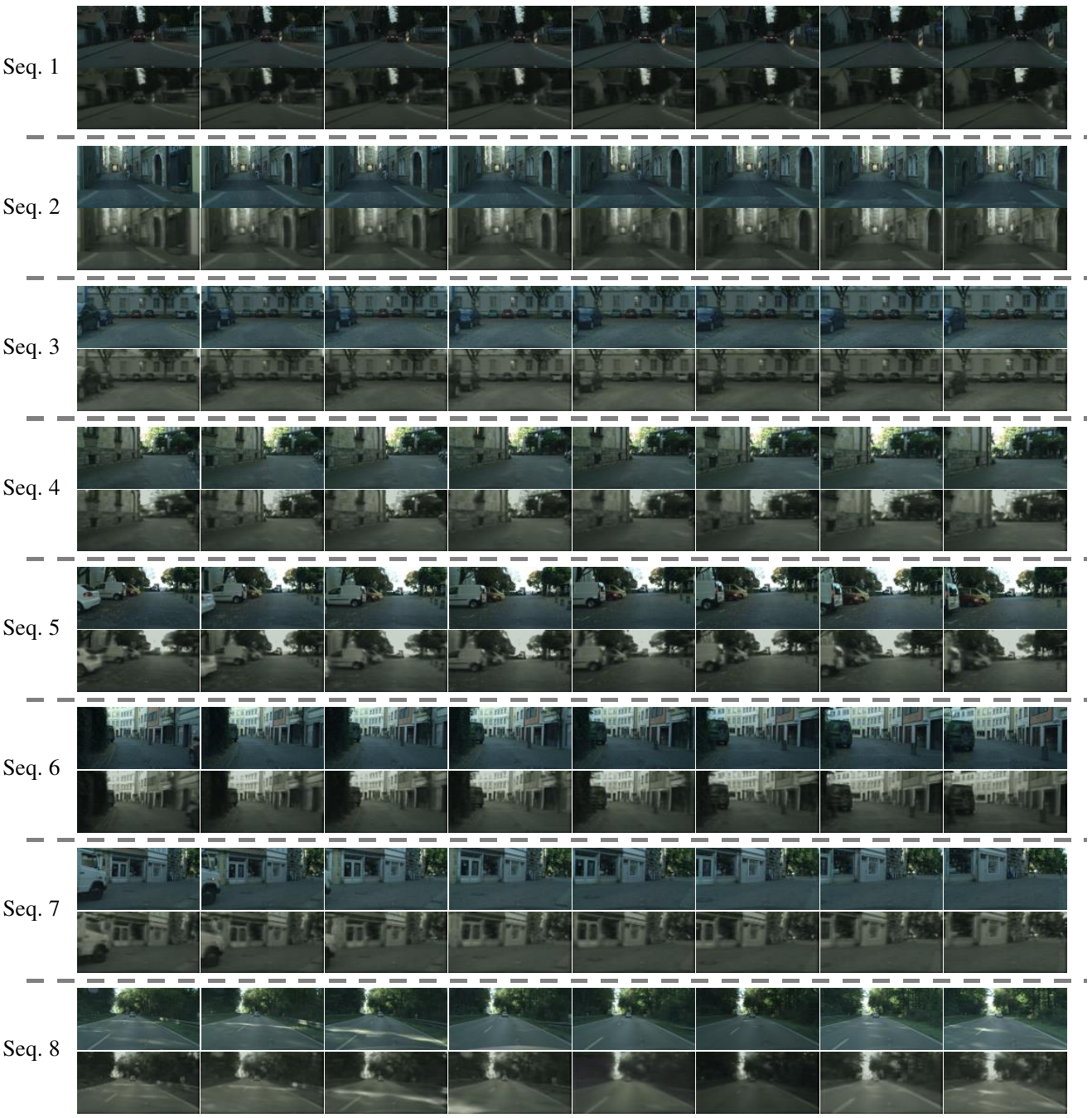}
	\caption{Eight sequences of videos in Cityscapes val set with frame
          prediction results. For each sequence, the upper row contains eight
          ground truth frames and the bottom row contains frame predictions
          produced by FPNet in PEARL. It is observed that FPNet is able to model
          the structures of objects and stuff as well as the motion information
          of moving objects in videos. Best viewed in color and zoomed pdf.}
 	\label{fig:supp_v2f_vis}
      \end{figure*}

Please refer to Figure~\ref{fig:supp_v2f_vis};
\subsection{More Results of Video Scene Parsing}
Please refer to Figure~\ref{fig:supp_parse_vis}.

\begin{figure*}[t]
  \captionsetup[subfigure]{labelformat=empty}
  \centering
  \subfloat[]{
    \includegraphics[width=\linewidth]{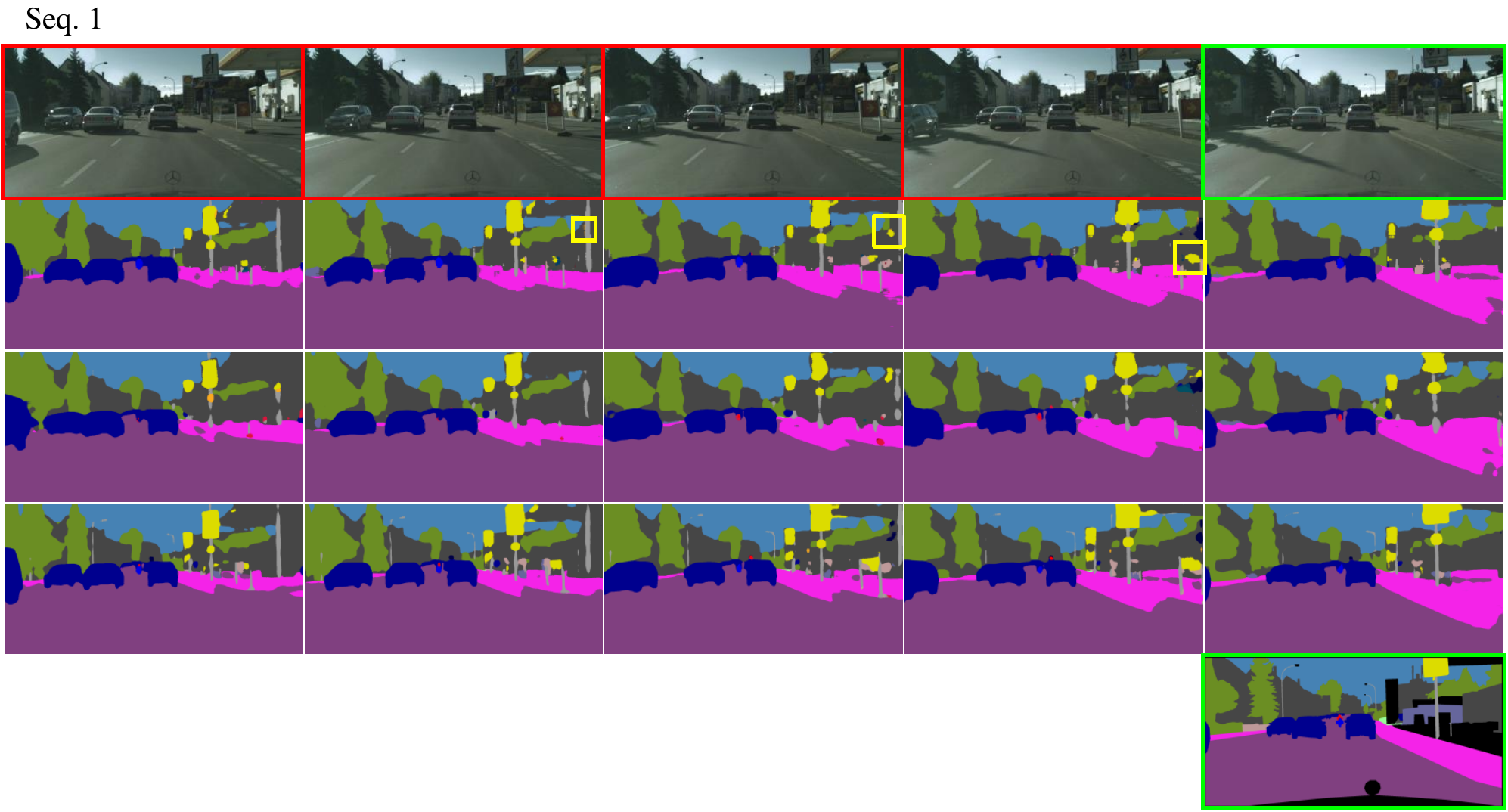}%
  }
  \phantomcaption\vspace{-5mm}\\
  \subfloat[]{
    \includegraphics[width=\linewidth]{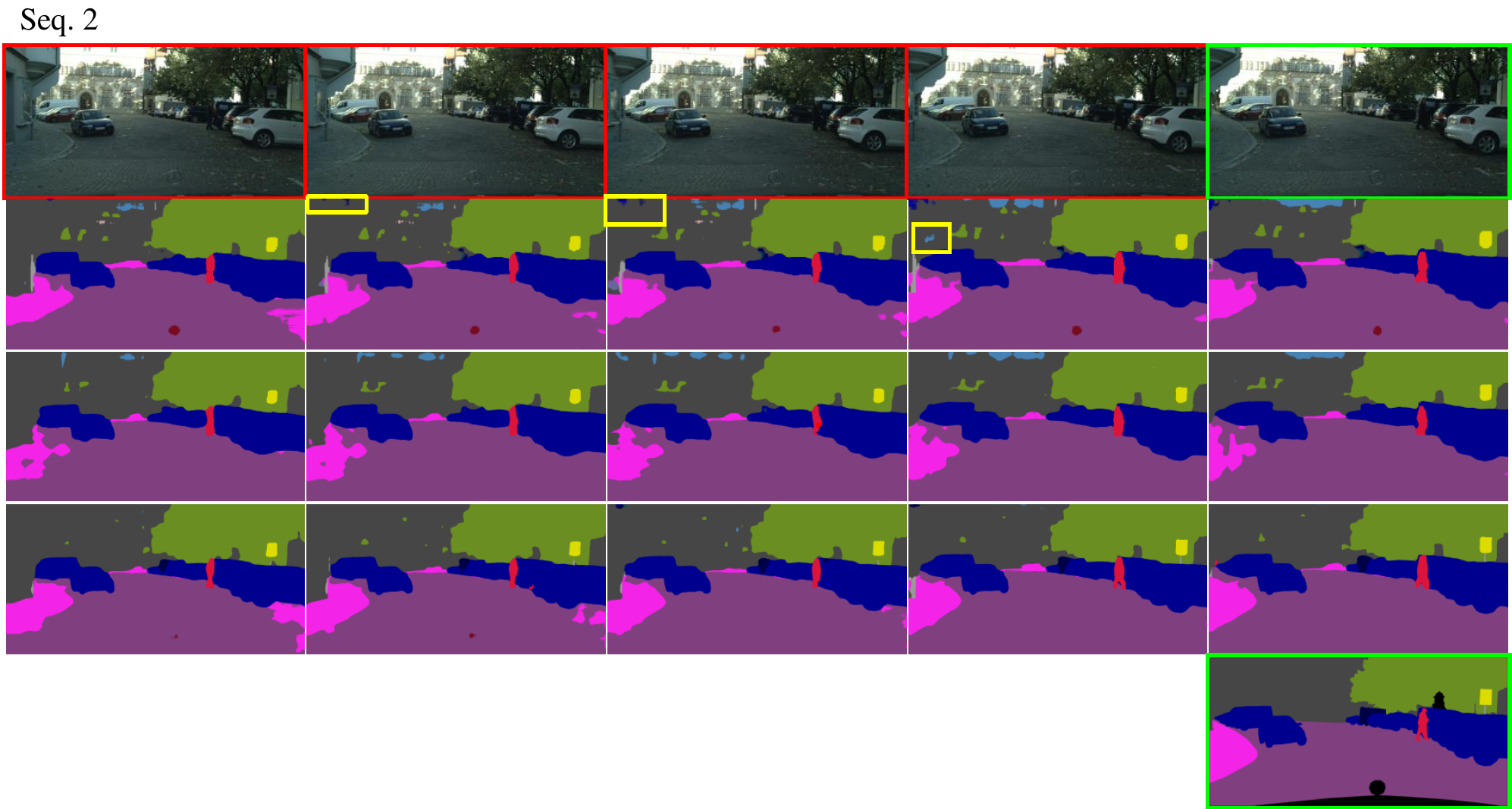}%
  }
  \phantomcaption

\end{figure*}

\begin{figure*}[t]
  \captionsetup[subfigure]{labelformat=empty}
  \centering
  \ContinuedFloat
  \subfloat[]{
    \includegraphics[width=\linewidth]{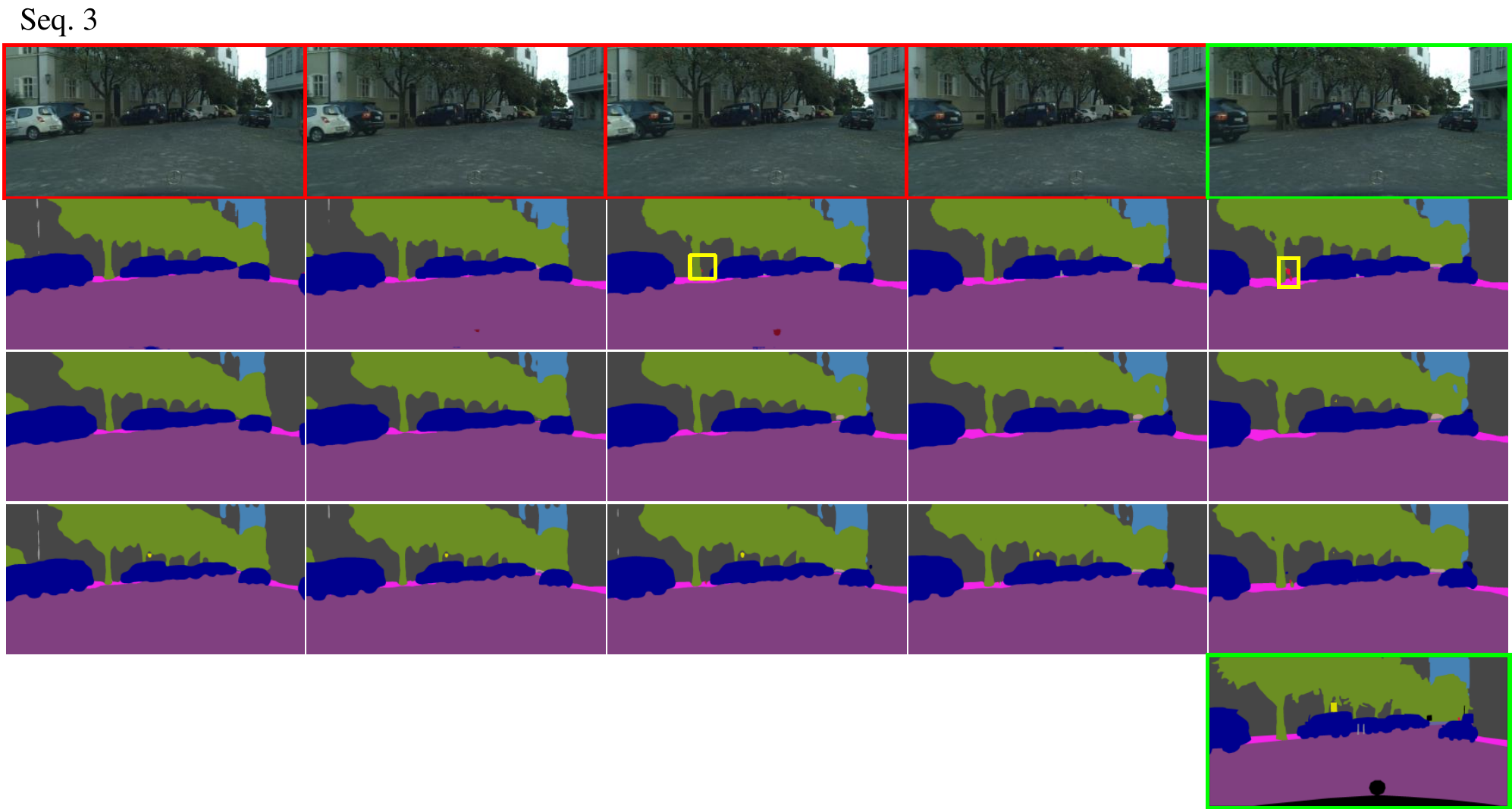}%
  }
  \phantomcaption\vspace{-5mm}\\
  \subfloat[]{
    \includegraphics[width=\linewidth]{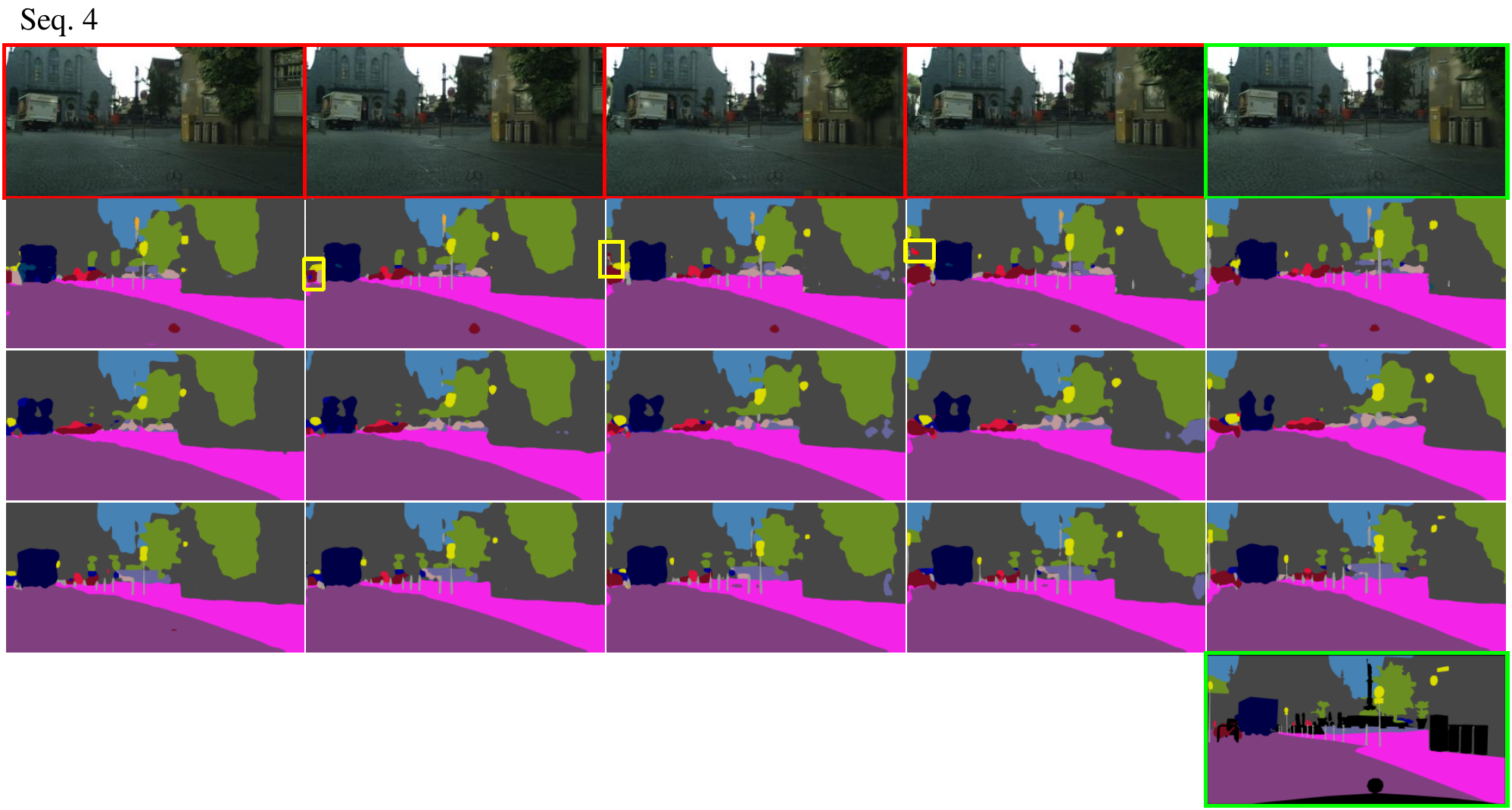}%
  }
  \phantomcaption

\end{figure*}

\begin{figure*}[t]
\captionsetup[subfigure]{labelformat=empty}
  \centering
  \ContinuedFloat
  \subfloat[]{
    \includegraphics[width=\linewidth]{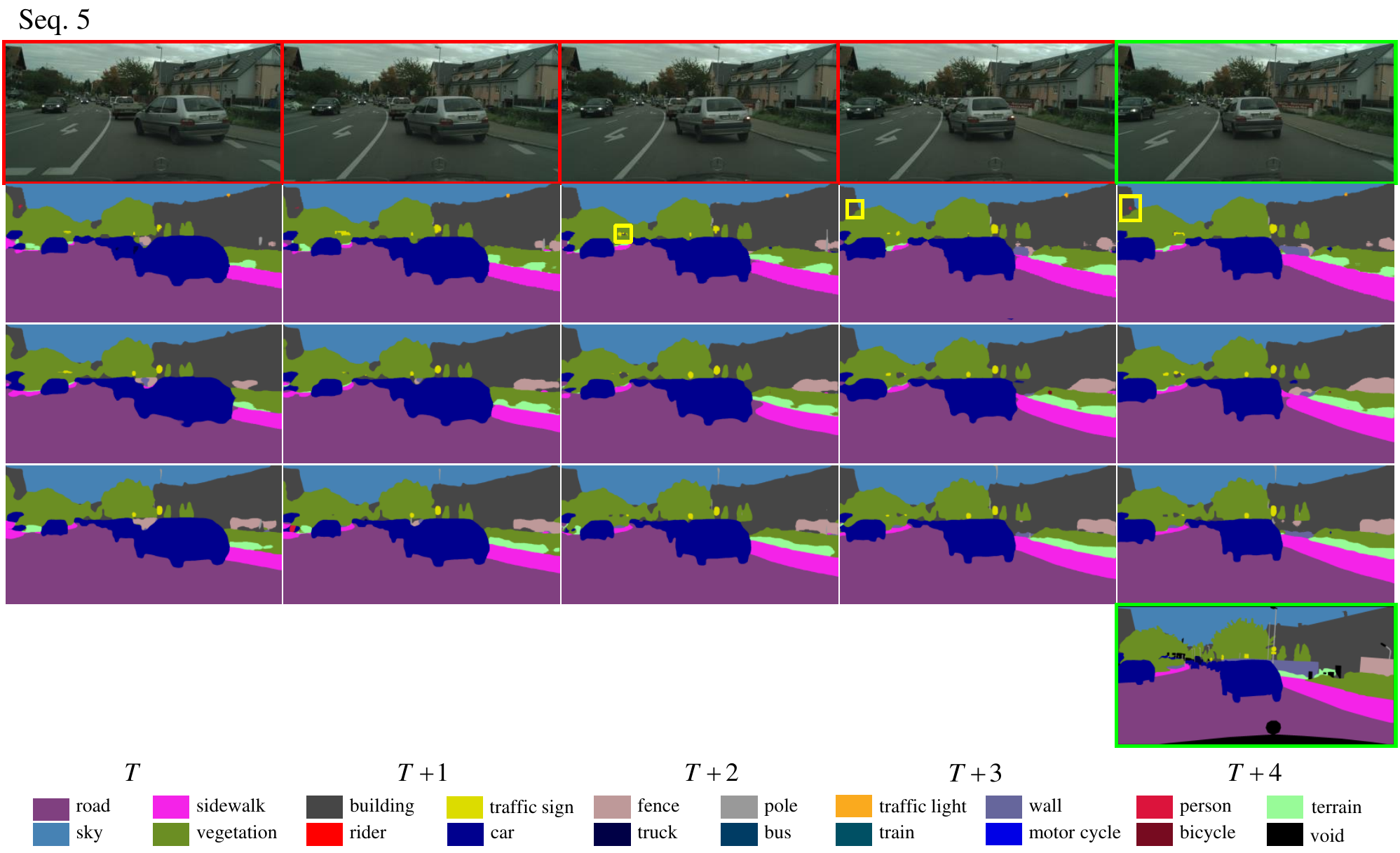}
  }\vspace{-2mm}
  \caption{Examples of parsing results of PEARL on Cityscape val set. All
    parsing maps have the same resolution as input frames.  \textbf{Top row}: a
    five-frame sequence (the four preceding frames are highlighted by red and
    the target frame to parse has a green boundary). \textbf{Second row}: frame
    parsing maps produced by the VGG16-baseline model. Since it cannot model
    temporal context, the baseline model produces parsing results with undesired
    inconsistency across frames as in yellow boxes.  \textbf{Third row}:
    predictive parsing maps output by PEARL. The inconsistent parsing regions in
    the second row are classified consistently across frames. \textbf{Fourth
      row}: parsing maps produced by PEARL with better accuracy and temporal
    consistency due to the combination of advantages of traditional image
    parsing model (the second row) and predictive parsing model (the third
    row). \textbf{Bottom row}: the ground truth label map (with green boundary)
    for the frame $T$+4. Best viewed in color and zoomed pdf.}
 \label{fig:supp_parse_vis}
\end{figure*}

\end{appendices}

\end{document}